\documentclass[a4paper, 10pt, conference]{ieeeconf}      

\IEEEoverridecommandlockouts                              

\usepackage{amsmath}
\usepackage{amssymb}
\usepackage{amsfonts}
\usepackage{latexsym}
\usepackage{lipsum}
\usepackage{multicol}
\usepackage{accents}
\usepackage{harpoon}
\usepackage{wrapfig}
\usepackage{tabularx}
\usepackage{euscript}
\usepackage{color}
\usepackage{xcolor}
\usepackage{graphicx}
\usepackage{float}
\usepackage[caption = false]{subfig}	
\usepackage{epsfig}
\usepackage{hyperref}
\usepackage{setspace}
\usepackage{fancyhdr}
\usepackage{textcomp}
\usepackage{cite}
\usepackage{epstopdf}
\usepackage{subfig}
\usepackage{verbatim}
\usepackage[left=0.806in,top=1in,right=0.806in,bottom=1in]{geometry} 
\DeclareGraphicsExtensions{.pdf,.eps,.jpg}

\usepackage{pifont}
\usepackage{latexsym}
\usepackage{psfrag}
\usepackage{stmaryrd}

\usepackage[linesnumbered,ruled,vlined]{algorithm2e}

\title{\LARGE \bf
Autonomous UAV Navigation Using Reinforcement Learning
}

\author{Huy X. Pham, Hung. M. La, David Feil-Seifer, Luan V. Nguyen
\thanks{Huy Pham and Luan Nguyen are PhD students, and Dr. Hung La is the director of the Advanced Robotics and Automation
(ARA) Laboratory. Dr. David Feil-Seifer is an Assistant Professor of Department of Computer Science and Engineering, University
of Nevada, Reno, NV 89557, USA. Corresponding author: Hung La, email: {\tt\small hla@unr.edu}}
}

\begin{document}
\maketitle
\thispagestyle{empty}
\pagestyle{empty}

\begin{abstract}
Unmanned aerial vehicles (UAV) are commonly used for missions in unknown environments, where an exact mathematical model of the environment may not be available. This paper provides a framework for using reinforcement learning to allow the UAV to navigate successfully in such environments. We conducted our simulation and real implementation to show how the UAVs can successfully learn to navigate through an unknown environment. Technical aspects regarding to applying reinforcement learning algorithm to a UAV system and UAV flight control were also addressed. This will enable continuing research using a UAV with learning capabilities in more important applications, such as wildfire monitoring, or search and rescue missions. 
\end{abstract}


\section{Introduction}\label{S.intro}
Using unmanned aerial vehicles (UAV), or drones, in missions involving navigating through unknown environment, such as wildfire monitoring~\cite{pham2017distributed}, target tracking~\cite{woods2016novel, woods2015dynamic, munoz2017adaptive}, or search and rescue~\cite{tomic2012toward}, is becoming more widespread, as they can host a wide range of sensors to measure the environment with relative low operation costs and high flexibility. One issue is that most current research relies on the accuracy of the model describing the target, or prior knowledge of the environment~\cite{la2015multi, la2015cooperative}. It is, however, very difficult to attain this in most realistic implementations, since the knowledge and data regarding the environment are normally limited or unavailable. Using reinforcement learning (RL) is a good approach to overcome this issue because it allows a UAV or a UAV team to learn and navigate through the changing environment without a model of the environment~\cite{sutton1998reinforcement}.

RL algorithms have already been extensively researched in UAV applications, as in many other fields of robotics~\cite{la2015multirobot, la2013cooperative}. Many papers focus on applying RL algorithm into UAV control to achieve desired trajectory tracking/following. In~\cite{faust2013learning}, Faust et al. proposed a framework using RL in motion planning for UAV with suspended load to generate trajectories with minimal residual oscillations. Bou-Ammar et al.~\cite{bou2010controller} used RL algorithm with fitted value iteration to attain stable trajectories for UAV maneuvers comparable to model-based feedback linearization controller. A RL-based learning automata designed by Santos et al.~\cite{dos2012design} allowed parameters tuning for a PID controller for UAV in a tracking problem, even under adversary weather conditions. Waslander et al.~\cite{waslander2005multi} proposed a test-bed applying RL for accommodating the nonlinear disturbances caused by complex airflow in UAV control. Other papers discussed problems in improving RL performance in UAV application. Imanberdiyev et al.~\cite{imanberdiyev2016autonomous} used a platform named TEXPLORE which processed the action selection, model learning, and planning phase in parallel to reduce the computational time. Zhang et al.~\cite{zhang2015geometric} proposed a geometry-based Q-learning to extend the RL-based controller to incorporate the distance information in the learning, thus lessen the time needed for an UAV to reach a target.

However, to the best of our knowledge, there are not many papers discussing about using RL algorithm for UAVs in high-level context, such as navigation, monitoring or other complex task-based applications. Many papers often did not provide details on the practical aspects of implementation of the learning algorithm on physical UAV systems. In this paper, we provide a detailed implementation of a UAV that can learn to accomplish tasks in an unknown environment. Using a simple RL algorithm, the drone can navigate successfully from an arbitrary starting position to a goal position in shortest possible way. The main contribution of the paper is to provide a framework for applying a RL algorithm to enable UAV to operate in such environment. The remaining of the paper is organized as follows. Section II provides more detail on problem formulation, and the approach we use to solve the problem. Basics in RL and how we design the learning algorithm are discussed in section III. We conduct a simulation of our problem on section IV, and provide details on UAV control in section V. Subsequently, a comprehensive implementation of the algorithm will be discussed in section VI. Finally, we conclude our paper and provide future work in section VII.

\section{Problem Formulation}

\begin{figure}[htb]
\centering
\includegraphics[width=\columnwidth]{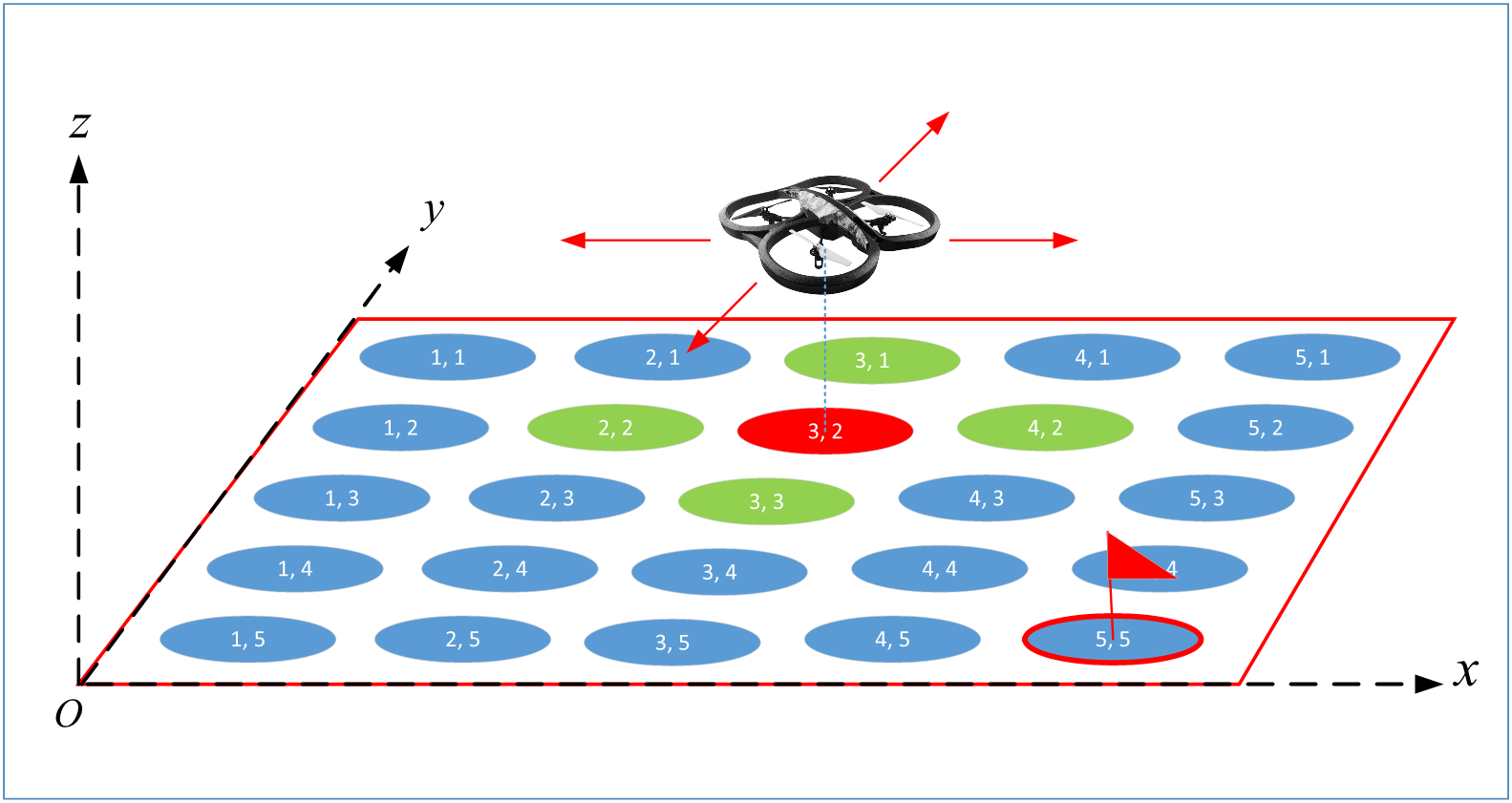}
  \caption{A UAV navigating in closed environment with discretized state space, represented by discrete circles. The red circle is the UAV's current state, the green circles are the options that the UAV can choose in the next iteration. The goal is marked by a red flag.}
  \label{F.Problem}
\vspace{0pt}
\end{figure}

Suppose that we have a closed environment in which the prior information about it is limited. We would like a flying robot, for example a quadcopter-type UAV, start at an arbitrary position to reach a goal that is pre-described to the robot (Figure \ref{F.Problem}). We assume that at any position, the UAV can observe its state, i.e. its position. If we have full information about the environment, for instance, the exact distance to the target or the locations of the obstacles, a robot motion planning can be constructed based on the model of the environment, and the problem becomes common. Traditional control methods, such as potential field~\cite{ge2002dynamic, woods2017novel}, are available to solve such problem. In many realistic cases, however, building models is not possible because the environment is insufficiently known, or the data of the environment is not available or difficult to obtain. Since RL algorithms can rely only on the data obtained directly from the system, it is a natural option to consider for our problem. In the learning process, the agent needs to map the situations it faces to appropriate actions so as to maximize a numerical signal, called reward, that measures the performance of the agent. 

\begin{figure}[htb]
\centering
\includegraphics[width=\columnwidth]{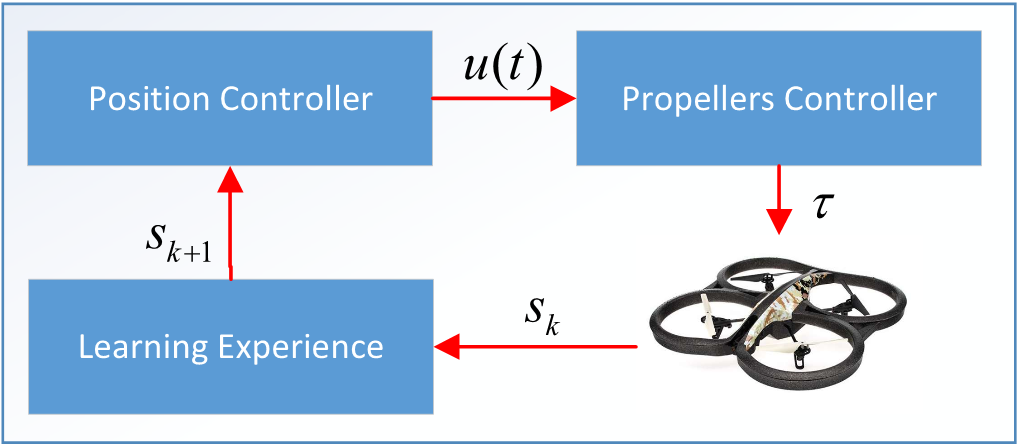}
  \caption{Reinforcement Learning model.}
  \label{F.DroneController}
\vspace{-0pt}
\end{figure}

To carry out the given task, the UAV must have a learning component to enable it to find the way to the goal in an optimal fashion. Based on its current state $s_{k}$ (e.g, UAV's position) and its learning model, the UAV decides the action to the next state $s_{k+1}$ it wants to be. A desired position then will be taken as input to the position controller, that calculates the control input $u(t)$ to a lower-level propellers controller. This low-level controller will control the motors of the UAV to generate thrust force $\tau$ to drive it to the desired position. Note that the position controller must be able to overcome the complex nonlinear dynamics of UAV system, to achieve stable trajectories for the UAV when flying, as well as hovering in the new state. Figure \ref{F.DroneController} shows the block diagram of our controller.

\section{Reinforcement Learning and Q Learning}

\begin{figure}[htb]
\centering
\includegraphics[width=\columnwidth]{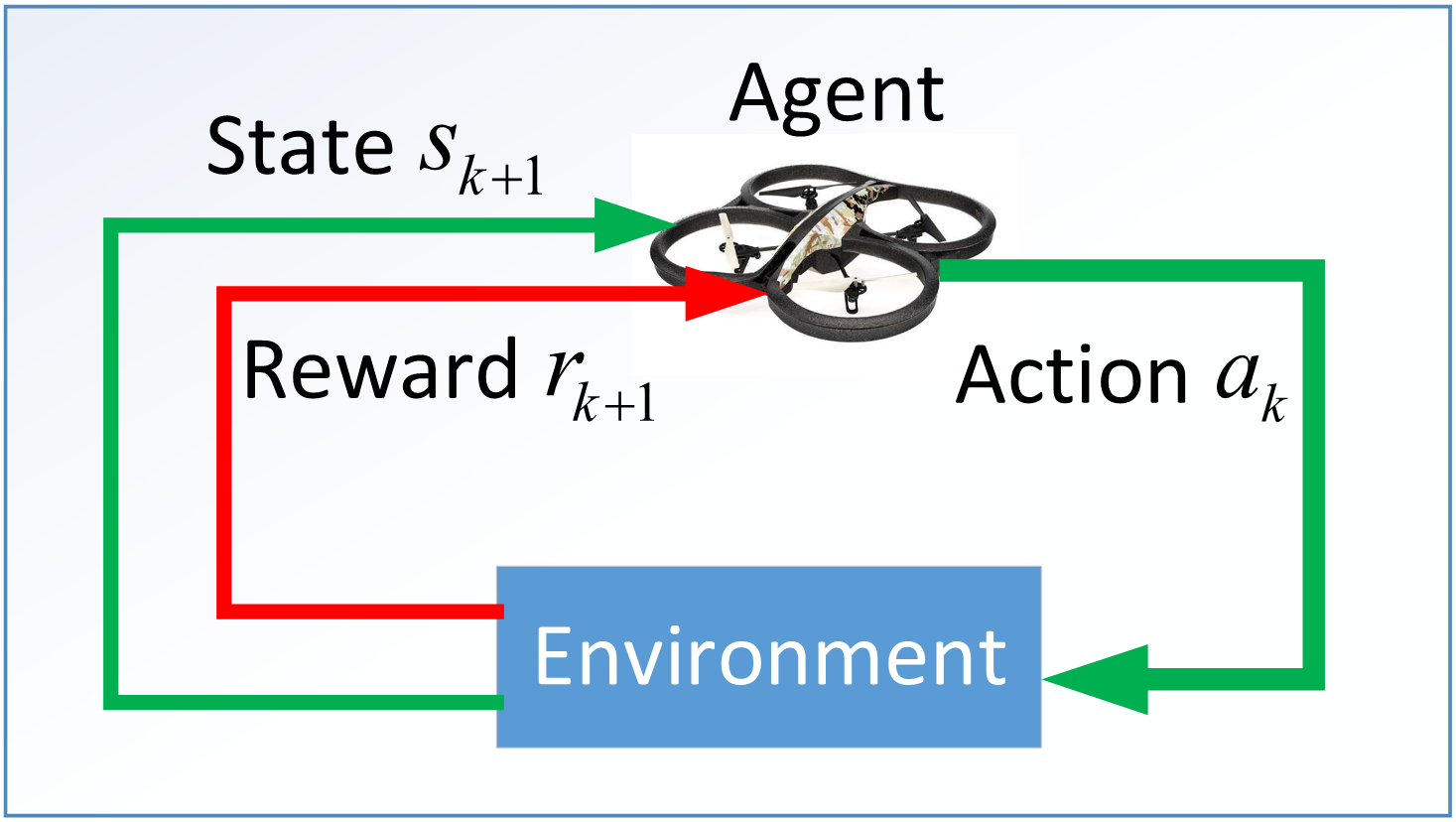}
  \caption{Reinforcement Learning model.}
  \label{F.RLmodel}
\vspace{-0pt}
\end{figure}

RL becomes popular recently thanks to its capabilities in solving learning problem without relying on a model of the environment. The learning model can be described as an agent\textendash environment interaction in Figure \ref{F.RLmodel}.  
An agent builds up its knowledge of the surrounding environment by accumulating its experience through interacting with the environment. Assuming that the environment has Markovian property, where the next state and reward of an agent only depends on the current state~\cite{sutton1998reinforcement}. The learning model can be generalized as a tuple $<S, A, T, R>$, where:

\begin{itemize}
	\item $S$ is a finite state list, and $s_{k} \in S$ is the state of the agent at step $k$;
	\item $A$ is a finite set of actions, and $a_{k} \in A$ is the action the agent takes at step $k$;
	\item $T$ is the transition probability function, $T:S \times A \times S  \rightarrow [0, 1]$, is the probability of agent that takes action $a_{k}$ to move from state $s_{k}$ to state $s_{k+1}$. In this paper, we consider our problem as a deterministic problem, so as $T(s_{k}, a_{k}, s_{k+1}) = 1$.
	\item $R$ is the reward function: $R: S \times A \rightarrow \mathbb{R}$ that specifies the immediate reward of the agent for getting to state $s_{k+1}$ from $s_{k}$ after taking action $a_{k}$. We have: $R(s_{k}, a_{k}) = r_{k+1}$.
\end{itemize}

The objective of the agent is to find a course of actions based on its states, called a policy, that ultimately maximizes its total amount of reward it receives over time. In each state, a state - action value function $Q(s_{k}, a_{k})$, that quantifies how good it is to choose an action in a given state, can be used for the agent to determine which action to take. The agent can iteratively compute the optimal value of this function, and from which derives an optimal policy. In this paper, we apply a popular RL algorithm known as Q-learning~\cite{watkins1992q}, in which the agent computes optimal value function and records them into a tabular database, called Q-table. This knowledge can be recalled to decide which action it would take to optimize its rewards over the learning episodes. For each iteration, the estimation of the optimal state - action value function is updated following the Bellman equation~\cite{sutton1998reinforcement}:
\begin{equation}\label{2.UpdateQindividual}
\begin{aligned}
	Q_{k+1}(s_{k}, a_{k})  &\leftarrow (1 - \alpha)Q_{k}(s_{k}, a_{k})\\ 
	&+ \alpha[ r_{k+1} + \gamma \underset{a\prime}{\max}Q_{k}(s_{k+1}, a\prime) ], \\
\end{aligned}
\end{equation}
\noindent

where $0 \leq \alpha \leq 0$ and $0 \leq \gamma \leq 0$ are learning rate and discount factor of the learning algorithm, respectively. To keep balance between exploration and exploitation actions, the paper uses a simple policy called $\epsilon$ greedy, with $0 < \epsilon < 1$, as follows:
\begin{equation}\label{2.epsilongreedy}
\begin{aligned}
	\ \pi(s) &=
	\begin{cases}
    	 \ \text{a random action } a, &  \text{with probability } \epsilon; \\
	\  a \in \underset{a\prime}{arg\max} Q_{k}(s_{k}, a\prime), & \text{otherwise.} \\
    	\end{cases}
\end{aligned}
\end{equation}

In order to use Q-learning algorithm, one must define the set of states $S$, actions $A$ and rewards $R$ for an agent in the system. Since the continuous space is too large to guarantee the convergence of the algorithm, in practice, normally these set will be represented as discrete finite sets approximately~\cite{busoniu2010reinforcement}. In this paper, we consider the environment as a finite set of spheres with equal radius $d$, and their centers form a grid. The center of the sphere now represents a discrete location of the environment, while the radius $d$ is the error deviation from the center. It is assumed that the UAV can generate these spheres for any unknown environment. The state of an UAV is then defined as their approximate position in the environment, $s_{k} \triangleq c = [x_{c}, y_{c}, z_{c}] \in S$, where $x_{c}$, $y_{c}$, $z_{c}$ are the coordinates of the center of a spheres $c$ at time step $k$. For simplicity, in this paper we will keep the altitude of the UAV as constant to reduce the number of states. The environment becomes a 2-D environment and the spheres now become circles. The state of the drone at time step $k$ is the lateral position of center $c$ of a circle $s_{k} = [x_{c}, y_{c}]$. Figure \ref{F.Problem} shows the discrete state space of the UAV used in this paper.

In each state, the UAV can take an action $a_{k}$ from a set of four possible actions $A$: heading North, West, South or East in lateral direction, while maintaining the same altitude. After an action is decided, the UAV will choose an adjacent circle where position is corresponding to the selected action. Note that the its new state $s_{k+1}$ is now associated with the center of the new circle. Figure \ref{F.Problem} shows number of options the UAV can take (in green color) in a particular state. Note that if the UAV stays in a state near the border of the environment, and selects an action that takes it out of the space, it should stay still in the current state. The rewards that an UAV can get depend whether it has reached the pre-described goal $G$, recognized by the UAV using a specific landmark, where it will get a big reward. Reaching other places that is not the desired goal will result in a small penalty (negative reward):
\begin{equation}\label{2.Reward}
\begin{aligned}
	\ r_{k+1} &= 
	\begin{cases}
	\  100, &  if  \ s_{k+1} \equiv G \\
    	 \ -1, &  otherwise. \\
    	\end{cases}
\end{aligned}
\end{equation}
\noindent

\section{Controller Design and Algorithm}

\begin{figure}[htb]
\centering
\includegraphics[width=\columnwidth]{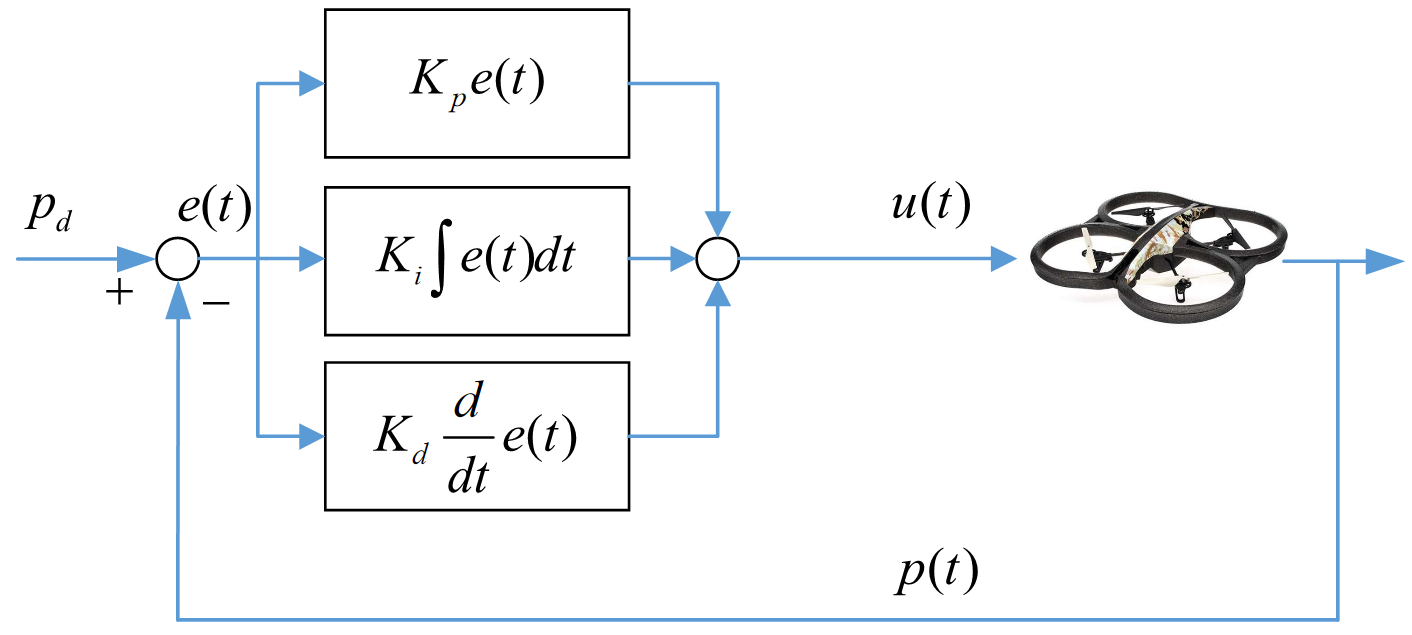}
  \caption{The PID control diagram with 3 components: proportional, integral and derivative terms.}
  \label{F.PIDController}
\vspace{-0pt}
\end{figure}

\begin{figure}[htb]
\centering
\includegraphics[width=\columnwidth, height=8cm]{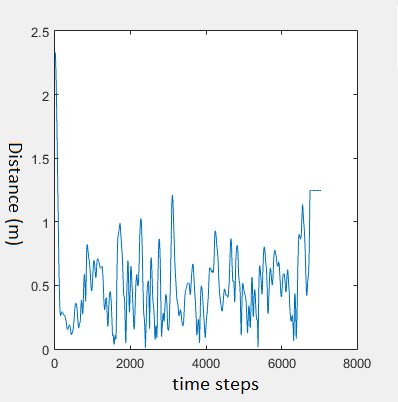}
  \caption{Distance error between the UAV and the target before tuning.}
  \label{F.PDbefore}
\vspace{-0pt}
\end{figure}

In this section, we provide a simple position controller design to help a quadrotor-type UAV to perform the action $a_{k}$ to translate from current location $s_{k}$ to new location $s_{k+1}$ and stay hovering over the new state within a small error radius $d$. Define $p_{t}$ is the real-time position of the UAV at time $t$, we start with a simple proportional gain controller: 
\begin{equation}\label{4.ProportionalControl}
\begin{aligned}
	u(t) = K_{p}(p(t) - s_{k+1}) = K_{p}e(t),
\end{aligned}
\end{equation}
\noindent
where $u(t)$ is the control input, $K_{p}$ is the proportional control gain, and $e(t)$ is the tracking error between real-time position $p(t)$ and desired location $s_{k+1}$. Because of the nonlinear dynamics of the quadrotor~\cite{woods2017novel}, we experienced excessive overshoots of the UAV when it navigates from one state to another (Figure \ref{F.PDbefore}), making the UAV unstable after reaching a state. To overcome this, we used a standard PID controller~\cite{dorf2011modern} (Figure \ref{F.PIDController}). Although the controller cannot effectively regulate the nonlinearity of the system, work such as~\cite{li2011dynamic, lee2012hovering} indicated that using PID controller could still yield relatively good stabilization during hovering.
\begin{equation}\label{4.ProportionalControl}
\begin{aligned}
	u(t) = K_{p}e(t) + K_{i}\int e(t)dt + K_{d}\frac{de}{dt}.
\end{aligned}
\end{equation}
\noindent

\begin{algorithm}
\DontPrintSemicolon 
\KwIn{Learning parameters: Discount factor $\gamma$, learning rate $\alpha$, number of episode $N$}
\KwIn{Control parameters: Control gains $K_{p}, K_{p}, K_{d}$, error radius $d$}
Initialize $Q_{0}(s,a) \leftarrow 0$, $\forall s_{0} \in S$, $\forall a_{0} \in A$;\\
\For{$episode =  1:N$}{
	Measure initial state $s_{0}$\\
	\For{$k =  0, 1, 2,...$}{
		Choose $a_{k}$ from $A$ using policy (\ref{2.epsilongreedy})\\
		Take action $a_{k}$ that leads to new state $s_{k+1}$:\\

		\For{$t =  0, 1, 2,...$}{
		\begin{equation*}
		\begin{aligned}
		u(t) = K_{p}e(t) + K_{i}\int e(t)dt + K_{d}\frac{de}{dt}\\
		\end{aligned}
		\end{equation*}
		} until $||p(t) - s_{k+1}|| \leq d$

		Observe immediate reward $r_{k+1}$\\
		Update:\\
		\begin{equation*}
		\begin{aligned}
		Q_{k+1}(s_{k}, a_{k})  &\leftarrow (1 - \alpha)Q_{k}(s_{k}, a_{k})\\ 
		&+ \alpha[ r_{k+1} + \gamma \underset{a\prime}{\max}Q_{k}(s_{k+1}, a\prime) ] \\
		\end{aligned}
		\end{equation*}
		\noindent
	} until $s_{k+1} \equiv G$
}
\caption{{\sc PID + Q-Learning.}}
\label{algo1}
\end{algorithm}

\begin{figure}[htb]
\centering
\includegraphics[width=\columnwidth]{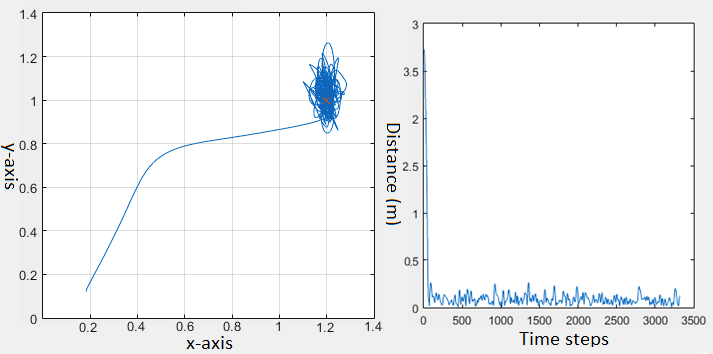}
  \caption{Distance error between the UAV and the target after tuning.}
  \label{F.PIDtuning}
\vspace{0 pt}
\end{figure}

\begin{figure}[htb]
\centering
\includegraphics[width=\columnwidth]{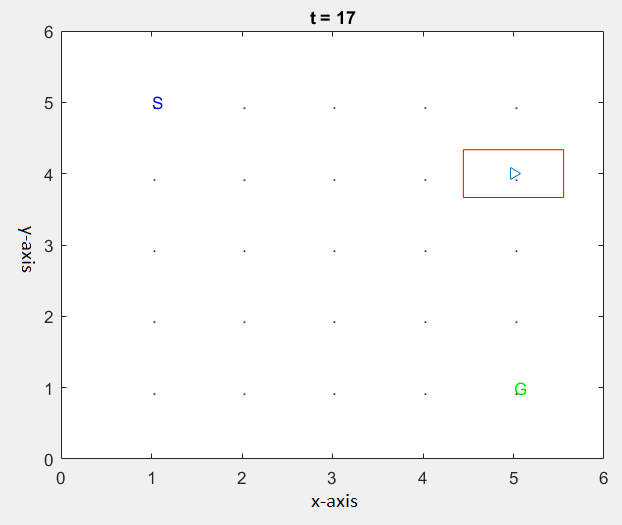}
  \caption{The simulated environment at time step $t = 17$. Label S shows the original starting point, and Label G shows the goal.}
  \label{F.Board}
\vspace{0pt}
\end{figure}

\begin{figure}[htb]
\centering
\includegraphics[width=\columnwidth]{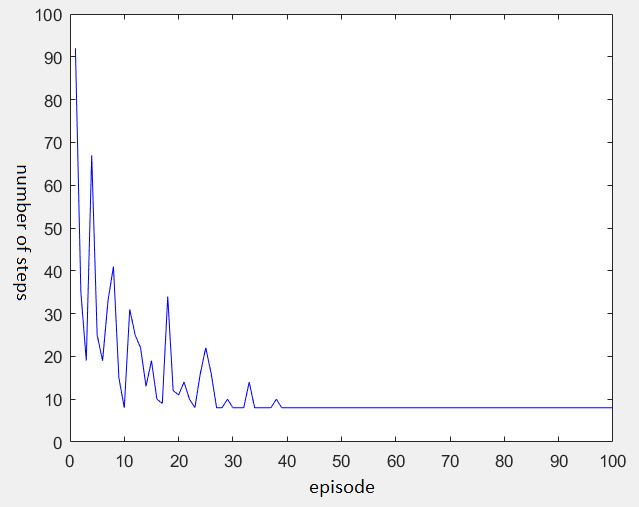}
  \caption{The time steps taken in each episode of the simulation.}
  \label{F.SimStep}
\vspace{0pt}
\end{figure}
\begin{figure}[htb!]
\centering
\includegraphics[width=\columnwidth]{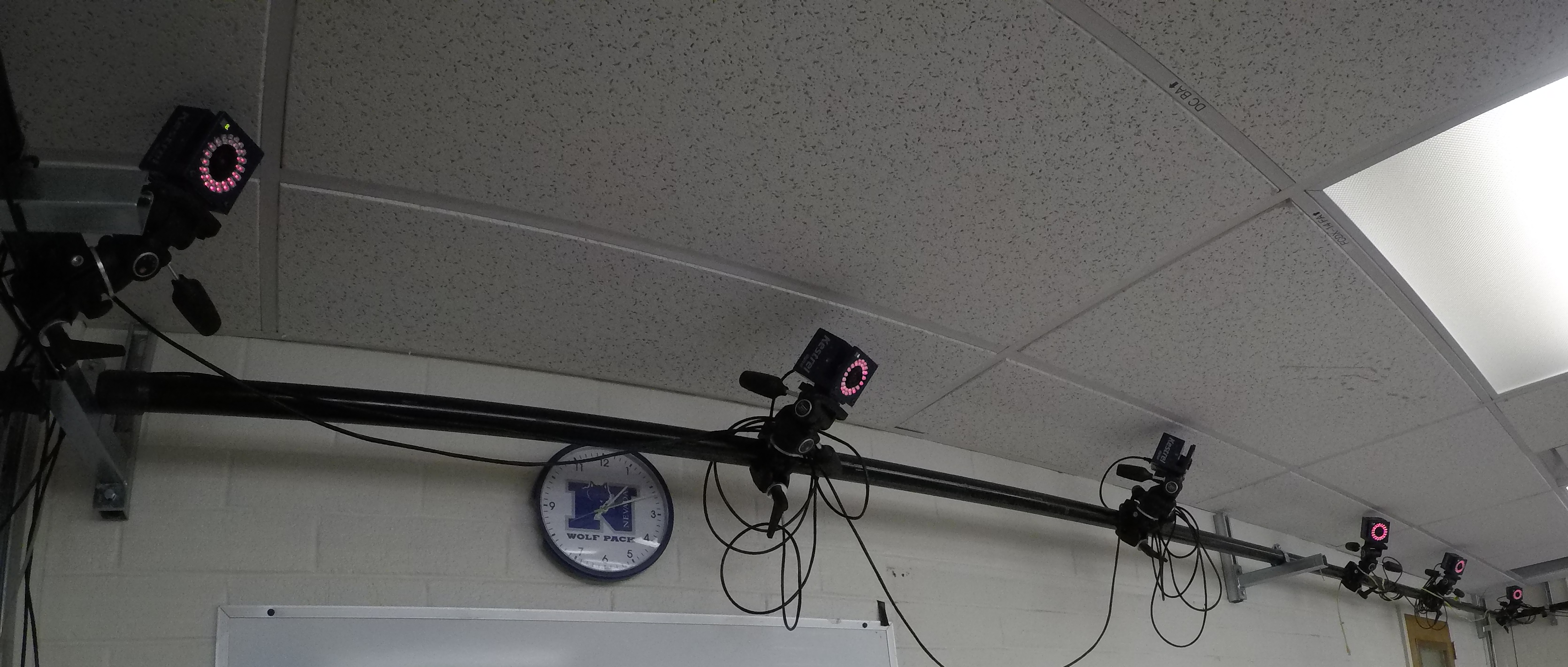}
  \caption{Motion capture system from Motion Analysis.}
  \label{F.Mocap}
\vspace{0pt}
\end{figure}

Generally, the derivative component can help decrease the overshoot and the settling time, while the integral component can help decrease the steady-state error, but can cause increasing overshoot. During the tuning process, we increased the Derivative gain while eliminated the Integral component of the PID control to achieve stable trajectory. Note that $u(t)$ is calculated in the Inertial frame, and should be transformed to the UAV's Body frame before feeding to the propellers controller as linear speed~\cite{woods2017novel}. Figure \ref{F.PIDtuning} shows the result after tuning. The UAV is now able to remain inside a radius of $d = 0.3$m from the desired state. The exact values of these parameters will be provided in section VI. Algorithm \ref{algo1} shows the PID + Q learning algorithm used in this paper.

\section{Simulation}

In this section, we conducted a simulation on MATLAB environment to prove the navigation concept using RL. We defined our environment as a 5 by 5 board (Figure \ref{F.Board}). Suppose that the altitude of the UAV was constant, it actually had 25 states, from $(1, 1)$ to $(5, 5)$. The UAV was expected to navigate from starting position at $(1, 1)$ to goal position at $(5, 5)$ in shortest possible way. Each UAV can take four possible actions to navigate: forward, backward, go left, go right. The UAV will have a big positive reward of +100 if it reaches the goal position, otherwise it will take a negative reward (penalty) of -1. We chose a learning rate $\alpha = 0.1$, and discount rate $\gamma = 0.9$.

Figure \ref{F.SimStep} shows the result of our simulation on MATLAB. It took 39 episodes to train the UAV to find out the optimal course of actions it should take to reach the target from a certain starting position. The optimal number of steps the UAV should take was 8 steps, resulting in reaching the target in shortest possible way.

\section{Implementation}

\begin{figure*}[t]
\centering
\includegraphics[width=\textwidth]{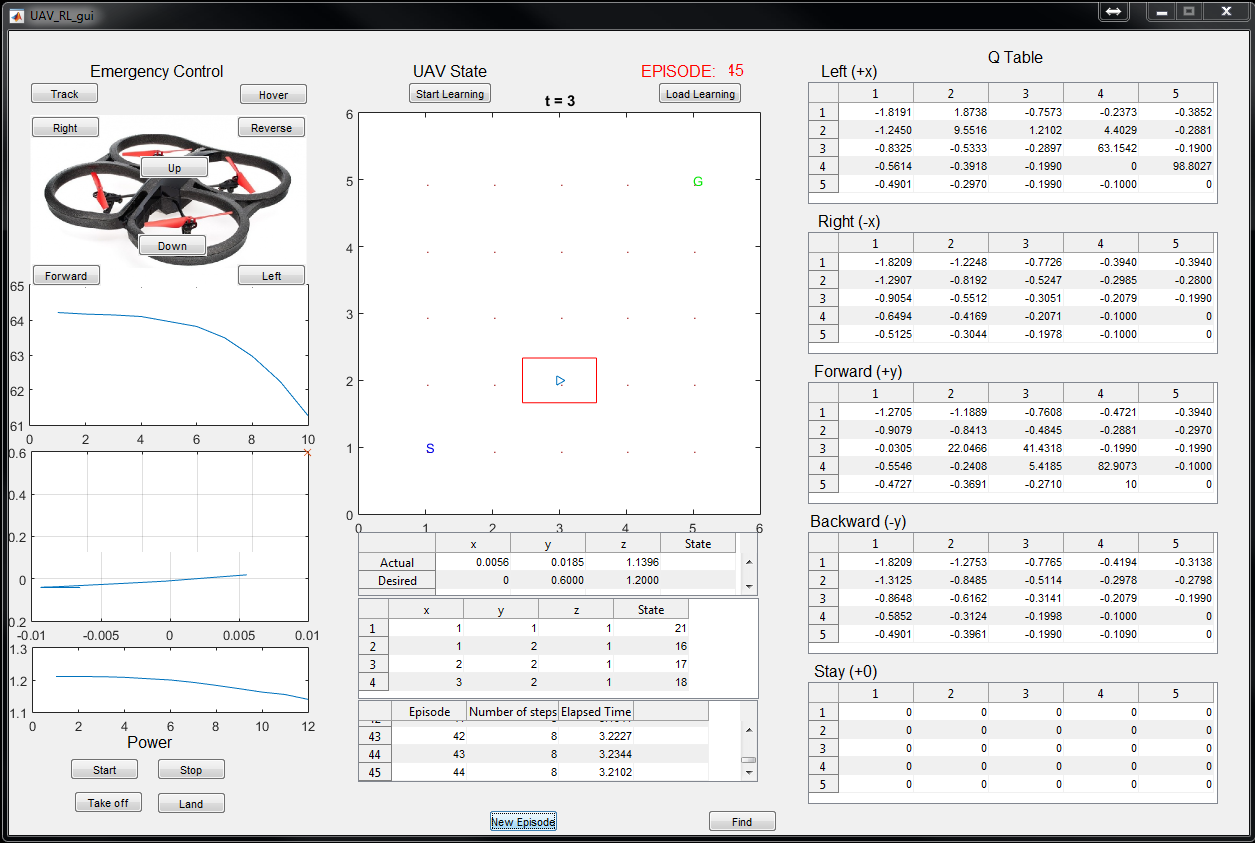}
  \caption{MATLAB GUI for control the learning process of the UAV.}
  \label{F.GUI}
\vspace{0pt}
\end{figure*}

\begin{figure}[htb]
\centering
\includegraphics[width=\columnwidth]{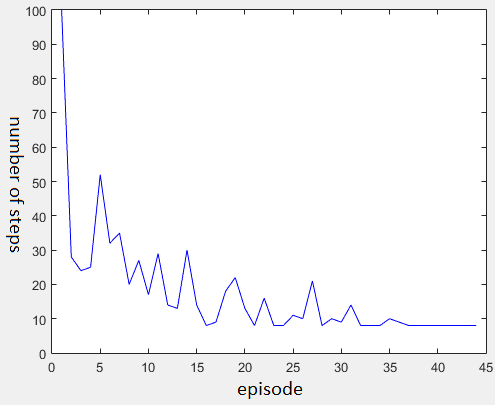}
  \caption{Number of steps in each episode in real implementation.}
  \label{F.Real}
\vspace{0pt}
\end{figure}

\begin{figure*}[htb]
 \centering
	\subfloat[t = 1]{\includegraphics[width=0.25\textwidth]{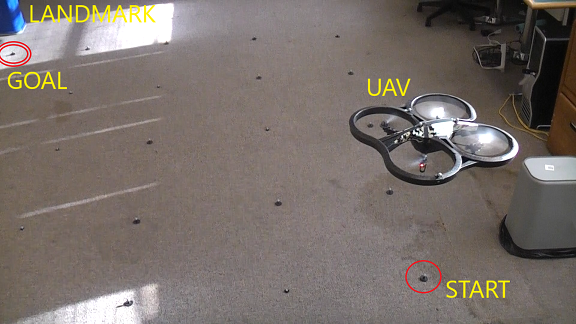}}
	\subfloat[t = 2]{\includegraphics[width=0.25\textwidth]{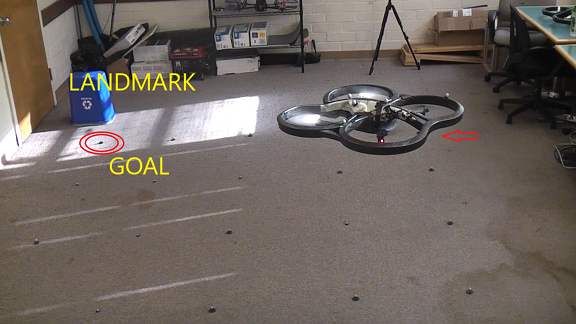}}
	\subfloat[t=  3]{\includegraphics[width=0.25\textwidth]{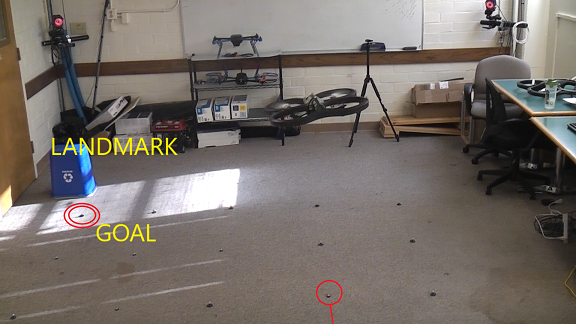}}
	\subfloat[t = 4]{\includegraphics[width=0.25\textwidth]{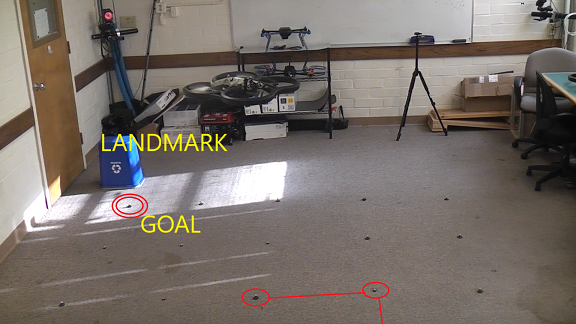}}
  \label{F.Last}
\vspace{0pt}
\end{figure*}
\begin{figure*}[htb]
 \centering
	\subfloat[t = 5]{\includegraphics[width=0.25\textwidth]{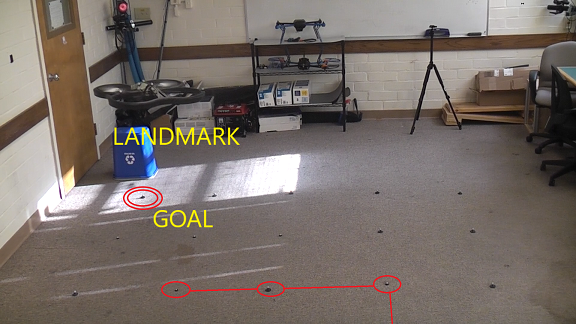}}
	\subfloat[t = 6]{\includegraphics[width=0.25\textwidth]{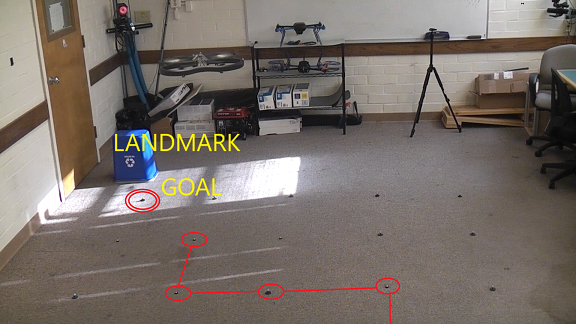}}
	\subfloat[t=  7]{\includegraphics[width=0.25\textwidth]{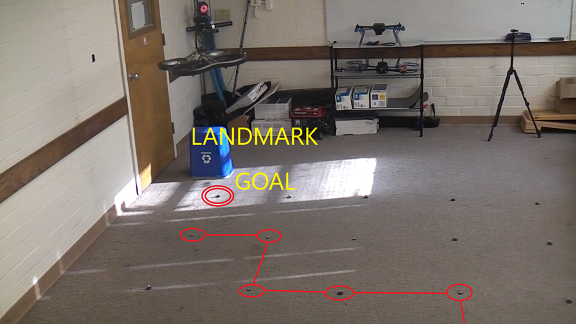}}
	\subfloat[t = 8]{\includegraphics[width=0.25\textwidth]{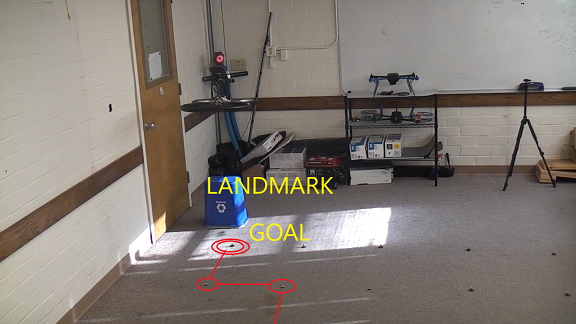}}
  \caption{Trajectory of the UAV during the last episode. It shows that the UAV reaches the target in the shortest possible way.}
  \label{F.Last}
\vspace{-0pt}
\end{figure*}

For our real-time implementation, we used a quadrotor Parrot AR Drone 2.0, and the Motion Capture System from Motion Analysis~\cite{motionanalysis} installed in our Advanced Robotics and Automation (ARA) lab (Figure \ref{F.Mocap}). The UAV could be controlled by altering the linear/angular speed, and the motion capture system provides the UAV's relative position inside the room. To carry out the algorithm, the UAV should be able to transit from one state to another, and stay there before taking new action. We implemented the PID controller in section IV to help the UAV carry out its action. Obviously, the learning process was a lengthy one. Therefore, to overcome the physical constraint on UAV's battery life cycle, we also designed a GUI on MATLAB to help discretize the learning process into episodes (Figure ~\ref{F.GUI}). For better control of the learning progress, the GUI shows information of the current position of the UAV within the environment, the steps the UAV has taken, the current values of Q table, and the result of this episode comparing to previous episodes. It also helped to save the data in case a UAV failure happened, allowing us to continue the learning progress after the disruption.

We carried out the experiment using identical parameters to the simulation. The UAV operated in a closed room, which is discretized as a 5 by 5 board. It did not have any knowledge of the environment, except that it knew when the goal is reached. Given that the altitude of the UAV was kept constant, the environment actually has 25 states. The objective for the UAV was to start from a starting position at $(1, 1)$ and navigate successfully to the goal state $(5, 5)$ in shortest way. Similar to the simulation, the UAV will have a big positive reward of +100 if it reaches the goal position, otherwise it will take a negative reward (penalty) of -1. For the learning part, we selected a learning rate $\alpha = 0.1$, and discount rate $\gamma = 0.9$. For the UAV's PID controller, the proportional gain $K_{p} = 0.8$, derivative gain $K_{d} = 0.9$, and integral gain $K_{i} = 0$. Similar to our simulation, it took the UAV 38 episodes to find out the optimal course of actions (8 steps) to reach to the goal from a certain starting position (Figure \ref{F.Real}). The difference between the first episode and the last ones was obvious: it took 100 steps for the UAV to reach the target in the first one, while it took only 8 steps in the last ones. Figure \ref{F.Last} shows the optimal trajectory of the UAV during the last episode.

\section{Conclusion}

This paper presented a technique to train a quadrotor to learn to navigate to the target point using a PID+ Q-learning algorithm in an unknown environment. The implementation and simulation outputs similar result, showed that the UAVs can successfully learn to navigate through the environment without the need of a mathematical model. This paper can serve as a simple framework for using RL to enable UAVs to work in an environment where its model is unavailable. For real-world deployment, we should consider stochastic learning model where uncertainties, such as wind and other dynamics of the environment, present in the system \cite{La_ICRA_2010,munoz2017adaptive}. In the future, we will also continue to work on using UAV with learning capabilities in more important application, such as wildfire monitoring, or search and rescue missions. The research can be extended into multi-agent systems \cite{LA_RAS2012, La_IEEE-C-2013}, where the learning capabilities can help the UAVs to have better coordination and effectiveness in solving real-world problem.

\addtolength{\textheight}{-12cm}   








\bibliographystyle{IEEEtran}
\bibliography{UAVsRL_bibliography}

\end{document}